\documentclass[letterpaper]{article}

\usepackage{natbib,alifeconf}  

\title{The Importance of Open-Endedness \\(for the Sake of
  Open-Endedness)}
\author{Tim Taylor$^{1,2,3}$\\
\mbox{}\\
$^1$Independent Researcher, Edinburgh, U.K.\\
$^2$Faculty of Information Technology, Monash University, Melbourne, Australia\\
$^3$International Programmes in Computing, University of London, U.K.\\
tim@tim-taylor.com}

\begin{document}
\maketitle

\section{Introduction}
A paper in the recent Artificial Life journal special issue on
open-ended evolution (OEE) presents a simple evolving computational
system that, it is claimed, satisfies all proposed requirements for
OEE \citep{Hintze:OpenEndedness}. Analysis and discussion of the
system are used to support the further claims that \emph{complexity} and
\emph{diversity} are the crucial features of open-endedness, and that we
should concentrate on providing proper definitions for those terms
rather than engaging in ``the quest for open-endedness for the sake of
open-endedness'' \citep[p.\ 205]{Hintze:OpenEndedness}. While I
wholeheartedly support the pursuit of precise definitions of
complexity and diversity in relation to OEE research, I emphatically
reject the suggestion that OEE is not a worthy research topic in its
own right. In the same issue 
of the journal, I presented a ``high-level conceptual 
framework to help orient the discussion and implementation of
open-endedness in evolutionary systems''
\citep{Taylor:Evolutionary}. In the current brief contribution I apply
my framework to Hinzte's model to understand its limitations.
In so doing, I demonstrate the importance of
studying open-endedness for the sake of open-endedness.

\section{A framework for understanding OEE}
From my initial forays into OEE \citep{Taylor:PhD} onward,
I have always viewed it as an umbrella
term, or high-level goal, that encompasses many interlinked topics.
Over the last eight years I have made several attempts at making explicit
these different facets of OEE and how they fit together
\citep{Taylor:Exploring,Taylor:Requirements,Taylor:Evolutionary}. The
first two of these are, I believe, liable to be misunderstood by some
readers because, for the most part, the discussion within them relates
to self-reproducing systems (e.g. ``Tierra-like'' systems)---but
this assumption of the specific problem situation of self-reproducing
systems was perhaps not sufficiently emphasized in the
papers. However, my most recent and expansive attempt, and the one I
am most comfortable with, does not suffer from this weakness because
it does not make the same assumption; it can be applied to any
evolutionary system whether it involves self-reproducing organisms
evolving under natural selection or agents that are selected and
reproduced using extrinsic mechanisms (e.g. fitness functions) 
\citep{Taylor:Evolutionary}.

The framework set out in \citep{Taylor:Evolutionary} attempts to
describe the general \emph{design requirements} for
open-endedness. The idea is that this will be useful both in guiding
the design and implementation of OEE systems, and also in categorizing
and comparing the OE potential of existing systems.
The framework comprises three interrelated components:\footnote{As
  discussed in \citep{Taylor:Evolutionary}, I acknowledge that the
  framework could be further improved, e.g.\ by incorporating
  population-level processes (such as drift and neutral networks) and by
  adopting a more sophisticated treatment of the relationship
  between form, dynamics and behavior.}
\begin{enumerate}
\item The distinction between \emph{exploratory},
  \emph{expansive} and \emph{transformational} novelties. This
  is defined formally in \citep{Taylor:Evolutionary} but
  can be loosely thought of as the extent to which novelties
  are of the ``more of the same'' variety versus more fundamental and
  unexpected innovations.
\item A formalism of the basic processes required of any
  evolutionary system, cast as processes of \emph{generation} of
  phenotype from genotype,
  \emph{evaluation} of the phenotype, and \emph{reproduction} (with
  variation) of the phenotype. The formalism makes 
  explicit various influences and interactions between each of these
  processes, mediated by the laws of dynamics of the system and the
  biotic and abiotic context in which they occur.
\item The distinction between \emph{intrinsic} and \emph{extrinsic}
  implementations of each of these processes. Intrinsic processes
  (i.e.\ those explicitly implemented within the system itself)
  can evolve, whereas those implemented extrinsically
  to the system (e.g.\ external fitness functions) cannot.
  The greater the extent to which all three evolutionary
  processes are implemented intrinsically, the more deserving are the
  agents of the label \emph{self}-reproducing.
\end{enumerate}

A brief discussion of how the framework could be used to categorize
and compare the OE potential of existing systems was presented in
\citep[p.\ 220]{Taylor:Evolutionary}. I extend that discussion here by
applying the framework to the model proposed by Hintze.

\section{Hintze's model}

\subsubsection{Overview}
A simple evolving computational system is presented in
\citep{Hintze:OpenEndedness}. The model comprises a population of
agents in a discrete 2D space. Each agent starts life at the
center of the space and follows a trajectory defined by its
genome. The genome is a sequence of the symbols \emph{right},
\emph{left} and \emph{forward}. Agents only interact indirectly
through their shared trajectories; the fitness
function considers how many agents
traversed each square in the space, and awards points to each agent
that traversed a given square in inverse proportion to how many
other agents traversed the same square. Hintze finds that the
complexity of the agents' genomes (as defined by the Zlib compression
size) increases exponentially, as does the diversity \emph{between}
runs over generations (as defined by the mean Levenshtein distance
between all pairwise comparisons of a single randomly-chosen sequence
from each experiment). Hintze claims that his model fulfils all of
the hypothesized requirements for OEE suggested in several previous
publications, including \citep{Taylor:Requirements}.\footnote{I
  dispute this claim because I believe Hintze has misinterpreted some
  of the requirements in \citep{Taylor:Requirements}. However,
  this is incidental to the main point of the current contribution.}
Noting that the observed evolution of agent behavior is nevertheless
underwhelming, he argues that the results indicate the need for
better definitions of the types of complexity and diversity growth
required of an open-ended evolutionary system, and suggests that
``with the proper definitions of complexity and diversity,
open-endedness might be a natural consequence of these systems''
\citep[p.\ 204]{Hintze:OpenEndedness}.

\subsubsection{Critique}
Hintze asserts that ``the exciting property of an evolving system is
not its openness but instead the complexity of the actual evolved
solutions'' \citep[p.\ 200]{Hintze:OpenEndedness}. This is a highly
contentious assertion. Of course, evolutionary systems that can evolve
complex agents are of great interest in ALife. But following
Hintze's suggestion of concentrating on precise definitions of what
counts as ``interesting enough'' complexity and diversity in the
context of a given study is a very different research goal to
the study of OEE. \emph{Tierra} produced complex and diverse agents in
the initial generations, but after a while no further significant
innovations were observed \cite[p.\ 418]{Taylor:OEE}. It is
\emph{that} result---the lack of ongoing innovation in Tierra and
other computational evolutionary systems---that catalysed the
emergence of the field of OEE research. The pursuit of a defined
threshold level of complexity in an evolutionary system is a very
different goal to the pursuit of ongoing innovation. \emph{Both} are
perfectly valid and interesting research goals but they are different
goals.

In presenting my framework I argued that OEE ``comprises just two
essential processes: the ongoing exploration of a phenotype space
\ldots\ and the discovery of door-opening states in that space that
open up an expanded phenotype space'' \citep[p.\
222]{Taylor:Evolutionary}. An essential goal for OEE research is to
understand the \emph{mechanisms} by which these processes can be
implemented. In the original paper I discussed how the first of these could be
achieved ``by allowing for intrinsic means for ongoing modification of
[the processes of generation, evaluation and reproduction]'' \citep[p.\
216]{Taylor:Evolutionary}. Hintze's model only allows for ongoing
modification of one of these processes: \emph{evaluation}. It does
so by allowing a parameter of the evaluation function to change (the
biotic context of other agents in the population), but it does not allow
for the evaluation function itself to be changed (this is hard-coded
and extrinsic). The \emph{generation} of phenotype from genotype, and the
genetic operators involved in \emph{reproduction}, are also hard-coded
extrinsic processes. The model 
therefore has some limited capacity for ongoing exploration of phenotype
space but only via one of the possible mechanisms. By concentrating
on definitions of complexity and diversity, Hintze ignores
discussion of the one key aspect of his model that enables its
(limited) capacity for open-endedness---the role of the biotic context
in the evaluation function.

Furthermore, Hintze's model is completely lacking in the second
essential process of OEE---the ability to discover door-opening states
leading to expanded phenotype spaces.
It is therefore a model of \emph{exploratory} open-endedness only and
is incapable of producing \emph{expansive} or \emph{transformational}
innovations. The inability to discover expanded phenotype spaces
arises partially owing to the fixed one-to-one
relationship between genes and their meaning in the model (i.e.\ the
actions \emph{right}, \emph{left} and \emph{forward}). This is due to
the impoverished dynamics of the world which lacks any
laws of physics or possibility of action beyond what is directly
encoded in the genome. The genes in the model directly describe 
\emph{semantics}. In order for new semantics to 
evolve, the genes should describe \emph{syntactical} structures (or,
stated in physical terms, \emph{boundary conditions})
which interact with the laws of dynamics of the world, out of which
interactions semantics \emph{arise}.
This is the case in notable examples of interesting evolved agents such
as \citep{Sims:Evolving} and \citep{Baker:Emergent}, which both
involve agents evolving in simulated physical environments.
A preliminary discussion of these
issues was presented in \citep[pp. 221--222]{Taylor:Evolutionary},
although they deserve a more elaborate treatment in future work.

More can be said about the strengths and weaknesses of Hintze's
model in terms of its capacity for OEE, but the comments above at
least demonstrate that the concepts presented in
\citep{Taylor:Evolutionary} provide a useful framework upon which to
hang such a discussion.
OEE research seeks to understand the design of evolutionary
systems that exhibit an ongoing generation of creative innovations.
This is a different goal to studying the evolution of complexity or
diversity by themselves. Many interlinked topics must be assembled to
understand how to design and build OEE systems. It is only by studying
open-endedness, for the sake of open-endedness, that we might hope to
make progress towards this goal.



\footnotesize
\bibliographystyle{apalike}
\bibliography{taylor-importance-open-endedness}

\end{document}